\documentclass[sigconf]{acmart}

\AtBeginDocument{%
  }

\setcopyright{none}
\usepackage{multirow}
\usepackage{soul}
\usepackage[utf8]{inputenc}
\usepackage{geometry}
\usepackage{tabularx}
\usepackage{booktabs}
\usepackage{enumitem}
\usepackage{ragged2e}
\usepackage[utf8]{inputenc}
\usepackage{geometry}
\usepackage{tabularx}
\usepackage{booktabs}
\usepackage{enumitem}
\usepackage{ragged2e}

\begin{document}


\title{Can Large Multimodal Models Inspect Buildings? A Hierarchical Benchmark for Structural Pathology Reasoning}


\author{Hui Zhong}
\affiliation{%
  \institution{Hong Kong University of Science and Technology (Guangzhou)}
  \city{Guangzhou}
  \country{China}}
\email{hzhong638@connect.hkust-gz.edu.cn}
\orcid{0000-0002-4428-3554}

\author{Yichun Gao}
\affiliation{%
  \institution{Hong Kong University of Science and Technology (Guangzhou)}
  \city{Guangzhou}
  \country{China}}
\email{ygao514@connect.hkust-gz.edu.cn}

\author{Luyan Liu}
\affiliation{%
\institution{Hong Kong University of Science and Technology}
  \city{Hongkong}
  \country{China}}
\email{lliuct@connect.ust.hk}

\author{Hai Yang}
\affiliation{%
  \institution{Hong Kong University of Science and Technology}
  \city{Hong Kong}
  \country{China}
}

\author{Wang Wang}
\affiliation{%
  \institution{Hong Kong University}
  \city{Hong Kong}
  \country{China}
}

\author{Haowei Zhang}
\affiliation{%
  \institution{Hong Kong University}
  \city{Hong Kong}
  \country{China}
}
\authornote{Corresponding author.}

\author{Xinhu Zheng}
\affiliation{%
  \institution{Hong Kong University of Science and Technology (Guangzhou)}
  \city{Guangzhou}
  \country{China}}
\email{xinhuzheng@hkust-gz.edu.cn}
\authornotemark[1]

\renewcommand{\shortauthors}{Trovato et al.}

\begin{abstract}
Automated building facade inspection is a critical component of urban resilience and smart city maintenance. 
Traditionally, this field has relied on specialized discriminative models (e.g., YOLO, Mask R-CNN) that excel at pixel-level localization but are constrained to passive perception and worse generization without the visual understandng to interpret structural topology.
Large Multimodal Models (LMMs) promise a paradigm shift toward active reasoning, yet their application in such high-stakes engineering domains lacks rigorous evaluation standards.
To bridge this gap, we introduce a human-in-the-loop semi-automated annotation framework, leveraging expert-proposal verification to unify 12 fragmented datasets into a standardized, hierarchical ontology.
Building on this foundation, we present \textit{DefectBench}, the first multi-dimensional benchmark designed to interrogate LMMs beyond basic semantic recognition. 
\textit{DefectBench} evaluates 18 state-of-the-art (SOTA) LMMs across three escalating cognitive dimensions: Semantic Perception, Spatial Localization, and Generative Geometry Segmentation.
Extensive experiments reveal that while current LMMs demonstrate exceptional topological awareness and semantic understanding (effectively diagnosing "what" and "how"), they exhibit significant deficiencies in metric localization precision ("where"). 
Crucially, however, we validate the viability of zero-shot generative segmentation, showing that general-purpose foundation models can rival specialized supervised networks without domain-specific training. 
This work provides both a rigorous benchmarking standard and a high-quality open-source database, establishing a new baseline for the advancement of autonomous AI agents in civil engineering.
\end{abstract}







\maketitle

\section{Introduction}

The maintenance of aging urban building infrastructure is a critical challenge for public safety and smart city resilience \cite{zhang2025vision,xu2025image}.
Traditionally, this domain has been dominated by specialized discriminative models, such as YOLO, R-CNN, and VGG \cite{liu2022datasets,xu2025crack}.
While these models excel at \textit{passive perception} (i.e., localizing defects at the bounding-box level), they fundamentally lack the cognitive capability for active diagnosis.
Specifically, conventional discriminative approaches fail to capture high-order semantic relationships, thereby limiting current systems to rudimentary detection rather than actionable engineering diagnosis.
In contrast, Large Multimodal Models (LMMs) represent a paradigm shift towards \textit{active reasoning} \cite{mullappilly2024bimedix2}.
Empowered by open-world knowledge and zero-shot generalization, LMMs offer the unprecedented potential to evolve inspection systems from simple detectors into diagnostic agents capable of explaining risks and reasoning about structural integrity \cite{xu2025image, mohamed2025infragpt}.

However, the deployment of LMMs in this high-stakes vertical domain is hindered by two systemic barriers.
(1) \textit{Data Silos and Ontological Inconsistency}: The current landscape of building facade inspection is characterized by fragmentation \cite{liu2022datasets}. 
Existing public datasets function as isolated silos, exhibiting profound ontological discrepancies and semantic misalignments.
These inconsistencies manifest as conflicting taxonomies (e.g., disparate definitions for "spalling" vs. "peeling") and incompatible annotation granularities, ranging from coarse-grained classification to fine-grained pixel-level masks.
Furthermore, the ubiquitous absence of logic-intensive metadata across these disparate sources exacerbates data heterogeneity, thereby impeding the formation of a unified knowledge foundation necessary for evaluating the high-order reasoning of LMMs in complex engineering scenarios.
(2) \textit{Fragmentation of Evaluation Standards and Lack of Holistic Benchmarks}: While the civil engineering community has established benchmarks for specific discriminative tasks, such as crack classification or concrete spalling detection—these efforts remain narrowly confined to isolated sub-tasks and specific defect typologies \cite{liu2022datasets,zha2025dataset}.
Consequently, there is a systemic absence of a unified, multi-dimensional benchmarking framework capable of interrogating LMMs across the full cognitive spectrum. 
Existing standards lack the hierarchical structure necessary to evaluate the transition from coarse-grained semantic comprehension (Level 1) to spatial reasoning (Level 2) and fine-grained geometric quantification (Level 3). 
This structural gap leaves the "active reasoning" capabilities of LMMs unquantified in a comprehensive, integrated setting.

To address these systemic barriers, we synthesized a unified data landscape by consolidating 12 open-source building defect repositories through a rigorous pipeline of expert-led filtering and manual re-annotation.
This consolidation yielded a curated dataset of 1,488 high-resolution images, harmonized into a standardized taxonomy of four primary defect classes: crack, material loss, surface stain, and external fixings. Crucially, each entry is enriched with fine-grained sub-class labels and topology-aware metadata, explicitly encoding the spatial relationships and quantitative attributes of the defects to support high-order reasoning tasks.

Building upon this foundation, we establish \textit{DefectBench}, a hierarchical benchmark designed to rigorously interrogate LMMs in the context of building defect inspection. Unlike prior works that focus on isolated tasks, \textit{DefectBench} evaluates models across three escalating cognitive dimensions: Semantic Perception (Visual Understanding), Spatial Localization (Visual Grounding), and Generative Geometry Segmentation (Morphological Delineation)
We performed extensive experiments on 18 SOTA LMMs, including both open-source and proprietary models.
Through these assessments, we provide a standardized evaluation framework and a diverse collection of annotated datasets to bridge the existing gap between general-purpose LMM capabilities and specialized engineering requirements. Our primary contributions are summarized as follows:

\begin{itemize}
    \item \textbf{Unified Multi-Granularity Dataset}: We synthesized a harmonized repository from 12 heterogeneous data silos, offering a standardized ontology that spans from coarse-grained image tags to fine-grained pixel-level masks. This resource serves as a unified foundation for training and evaluating LMMs across classification, detection, and segmentation tasks.
    \item \textbf{Hierarchical Cognitive Benchmark:} We establish \textit{DefectBench}, the first multi-dimensional evaluation framework for facade inspection. By integrating Semantic Perception, Spatial Localization, and Generative Segmentation, our paradigm shifts the assessment standard from passive visual recognition to active diagnostic consistency.
    \item \textbf{Extensive Benchmarking and Analysis:} We conduct a comprehensive evaluation of 18 SOTA LMMs from semantic perception to topological reasoning. Our analysis provides the first systematic baseline for understanding the capabilities and limitations of zero-shot LMMs for building defect inspection.
    \item \textbf{Extensible Human-in-the-Loop Annotation Platform:} We introduce a semi-automated annotation toolkit engineered with a modular, plug-and-play architecture. By integrating an ensemble of specialized pre-trained models to generate candidate proposals for expert verification, this framework significantly accelerates high-precision data curation and facilitates the seamless integration of future other algorithms.
\end{itemize}


\smallskip
\noindent \textbf{Data and Code Availability:} 
To facilitate reproducibility and support future research in vertical domain inspection, the \textit{DefectBench} dataset and our semi-automated annotation toolkit will be released on a public repository upon the formal acceptance of this work.

\section{Related Work}

\subsection{From Visual Perception to Understanding}


Traditional research in building facade defect inspection has been predominantly task-specific and dataset-constrained, driven by the discriminative learning paradigm.
SOTA architectures, ranging from the real-time YOLO family to Transformer-based frameworks like DINO and Mask2Former, have achieved remarkable pixel-level precision in localized detection and segmentation \cite{liu2022datasets,wang2024text}.
However, these models are fundamentally restricted to passive perception, regarding defects as isolated geometric entities rather than interconnected structural anomalies. Thus they fail to interpret the underlying topological dependencies and causal relationships inherent in building deterioration.
In fact, facade anomalies exhibit a high degree of material-dependency and correlated deterioration mechanisms \cite{faqih2021defect,yan2025uav}.
For instance, material loss manifests as spalling in concrete but as peeling in plaster; furthermore, spalling is often accompanied by underlying cracks \cite{kottari2024bd3,zha2025dataset}.
Whereas a holistic structural health diagnosis necessitates visual contextualization, the ability to synthesize defect characteristics with domain-specific logical reasoning.
This is a critical challenge that traditional discriminative models fail to achieve due to their limited semantic depth and poor cross-domain generalization \cite{liu2022datasets,zha2025dataset}.

The emergence of LMMs provide a transformative solution, offering superior zero-shot generalization and visual-linguistic reasoning \cite{zhong2025can,feng2025verdi,zhang2025iiim}.
\cite{mohamed2025infragpt} proposed the InfraGPT, which attempts to bridge this gap by concatenating YOLO-based detection with the descriptive power of GPT-4o to generate maintenance action plans.
However, this pipeline-based approach tend to treat the LMM as a mere summarization layer rather than a core reasoning engine.
Despite these advancements, a critical evaluation vacuum persists: the community lacks a standardized, multi-dimensional benchmark to rigorously quantify the vertical domain intelligence of LMMs, specifically their capacity to navigate the complex logical and spatial constraints inherent in structural engineering\cite{chen2025chineseecomqa,zhang2025vision,xu2025image}.

\subsection{Building Defect Dataset and Benchmarks}

The advancement of automated structural health monitoring is intrinsically tied to the availability of high-quality annotated datasets.
Despite the proliferation of open-source data for infrastructure such as bridges and roadways \cite{dorafshan2018sdnet2018,ichi2021sdnet2021,liu2022datasets}, the domain of building facade inspection faces a persistent scarcity of comprehensive data.
This scarcity is primarily driven by the extreme intra-class variance in the shapes, sizes, textures, and visual manifestations of facade anomalies\cite{guo2024surface}.
There is a significant imbalance in defect representation with a vast majority of existing benchmarks focusing on a single, dominant category: cracks \cite{bai2020deep,liu2019deepcrack,dorafshan2018sdnet2018}.
While specialized datasets cater to crack-related classification, detection, and segmentation, other critical anomalies such as spalling, stains, and efflorescence remain underrepresented \cite{guo2024surface}.
These complex defects are largely confined to patch-level classification or bounding-box detection \cite{kottari2024bd3,zha2025dataset}, with a severe lack of high-fidelity masks required to support segmentation tasks\cite{zha2025dataset}.

Furthermore, the field is plagued by inconsistent annotation protocols and taxonomic ambiguity \cite{guo2021evaluation,guo2020faccade}.
The lack of a unified labeling standard exacerbates the complexity of data integration; for instance, plaster degradation is heterogeneously labeled as flake, peeling, or degraded plaster across different datasets \cite{mrpranavr_building_defect_app,j8j6-wp24-25}.
Moreover, the inherent cost and labor-intensity of manual expert annotation, coupled with scalability constraints, frequently lead to systemic issues such as label noise, missing annotations, and semantic misalignment\cite{guo2024surface}. 
These factors prevent the formation of robust datasets and benchmarks for structural health.
there is an urgent need for a unified, multi-level framework that bridges the gap between coarse-grained recognition and fine-grained segmentation across a holistic spectrum of defect types.
As LMMs increasingly permeate vertical domains, the absence of a standardized benchmark to quantify their domain-specific reasoning and diagnostic capabilities remains a critical void in the community.

To bridge these critical gaps, we introduce \textit{DefectBench}, a unified, multi-level framework designed to transcend the limitations of existing resources. 
Unlike previous datasets, DefectBench provides synchronized support for a hierarchy of tasks ranging from classification to semantic segmentation across a holistic spectrum of defect categories.
To empower advanced diagnostic reasoning, our dataset incorporates structured logical and spatial topological metadata, enabling the formulation of multi-level reasoning challenges. 
By establishing this benchmark, we provide a rigorous diagnostic environment to quantify the vertical domain intelligence of LMMs, facilitating their transition from passive visual recognition to active structural interpretation.

\section{DefectBench}

\subsection{Overview}
\begin{figure*}[ht]
  \centering
  \includegraphics[width=1\linewidth]{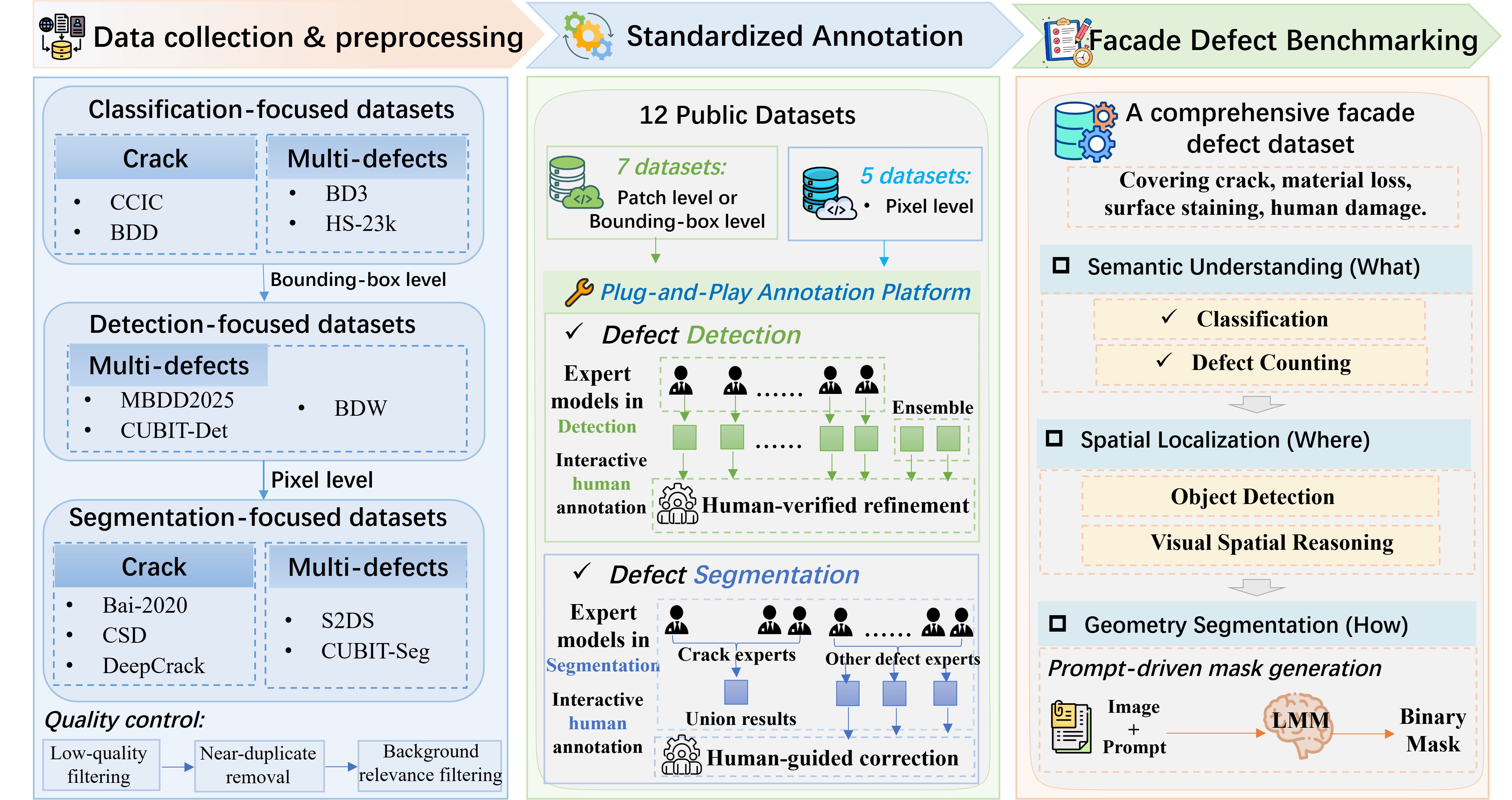}
  \caption{Overview of the DefectBench construction and evaluation framework.}
  \label{fig:overview}
\end{figure*}

In this section, we present DefectBench, a comprehensive and unified benchmark designed to elevate building facade inspection from simple visual recognition to complex multimodal reasoning.
As illustrated in Fig.\ref{fig:overview}, the construction of DefectBench follows a systematic three-stage pipeline: (1) Heterogeneous Data Integration, which curates and cleans raw data from 12 diverse sources; (2) Standardized Annotation, which employs a human-in-the-loop protocol to ensure semantic consistency; and (3) Hierarchical Task Formulation, which establishes a multi-level evaluative framework (What-Where-How) for LMMs.

\subsection{Heterogeneous Data Integration}
To ensure the robustness and taxonomic diversity of the benchmark, we aggregate data from 12 public datasets, aiming to cover more comprehensive spectrum of facade anomalies.
First, we systematically curate raw data based on their original primary focus, spanning three fundamental vision tasks, which ensures that \textit{DefectBench} captures a wide array of materials (e.g., concrete, plaster) and varied morphologies (e.g., cracks, material loss, staining).
For classification-focused data, we include single-category datasets like \textit{CCIC} \cite{bai2020deep} and \textit{BDD} \cite{mrpranavr_building_defect_app} for cracks, alongside multi-defect datasets such as \textit{BD3} \cite{kottari2024bd3} and \textit{HS-23k}\cite{j8j6-wp24-25}.
The detection-focused component primarily comprises multi-defect sources including \textit{MBDD2025} \cite{zha2025dataset}, \textit{BDW}\cite{heritage-defects_dataset}, and \textit{CUBIT-Det} \cite{11007070}.
Finally, the segmentation-focused subset aggregates pixel-level annotated data from \textit{Bai-2020}\cite{bai2020deep}, \textit{CSD}\cite{khanhha_crack_seg_dataset}, \textit{DeepCrack}\cite{liu2019deepcrack}, \textit{S2DS}\cite{benz2022image}, and \textit{CUBIT-Seg}\cite{yang2022datasets,11007070}.
Detailed statistics regarding image resolution, defect types, and dataset size for each source are summarized in Table \ref{tab:s1}.

To mitigate the "garbage in, garbage out" effect, we implement a three-stage curation pipeline to refine the raw data into high-quality benchmark samples.
Initially, we perform low-quality filtering by utilizing the Laplacian variance method to quantify image sharpness; samples identified as motion-blurred or possessing insufficient resolution are automatically discarded \cite{zhong2025predicting}.
To prevent evaluation bias caused by near-duplicate samples, we conduct Semantic De-duplication using a DINOv2 backbone to extract robust high-level features \cite{oquab2023dinov2}. 
By calculating cosine similarity across images, we identify and remove visually redundant data captured from slightly different angles. 
Lastly, through Contextual Relevance Filtering, we leverage CLIP to perform zero-shot semantic pruning \cite{radford2021learning}. 
This stage excludes noisy samples with non-building backgrounds or irrelevant clutter, ensuring the purity and domain-specificity of the structural dataset.

\subsection{Standardized Annotation Protocol}
To address the pervasive taxonomic ambiguity and inconsistent terminology in existing datasets, we implement a systematic annotation protocol. 
We first define a standardized, multi-level defect ontology to unify disparate labeling conventions. 
All anomalies are systematically categorized into four primary classes and eleven fine-grained sub-classes: (1) Crack, comprising linear cracks and map cracking; (2) Material Loss, including spalling, peeling, and structural bulging; (3) Surface Staining, covering rust stains, leakage marks, and corrosion; and (4) External Fixings, such as plant growth, graffiti, and surface contaminants. 
This hierarchical structure provides a consistent semantic foundation for cross-dataset integration and serves as the logic backbone for the subsequent multi-level reasoning tasks.

To facilitate rapid and precise semantic alignment, We design a Plug-and-Play Annotation Platform that utilizes a Human-in-the-loop strategy, architected into two specialized modules to handle distinct levels of visual granularity.
The Detection Refinement Module primarily addresses the spatial grounding of defects. 
For datasets with pre-existing annotations, the module supports human-verified refinement to correct potential box drifts. 
For unlabeled raw images, the platform integrates an ensemble of state-of-the-art (SOTA) detectors(e.g. YOLO12-M \cite{tian2025yolov12}, YOLO11-M, Faster R-CNN\cite{wu2019detectron2}, and RT-DETR\cite{lv2024rt}) to generate high-quality candidate proposals, which are then manually calibrated by experts to ensure rigorous spatial precision.
The Interactive Segmentation Module is specifically engineered to tackle the complex topologies of facade anomalies. 
Leveraging the refined bounding boxes as visual prompts, it employs SAM-3 for zero-shot mask generation, along with a domain-specific model zoo including SegFormer (b0/b4) \cite{xie2021segformer}, UNet-VGG16\cite{khanhha_crack_seg_dataset}, YOLOv8-crack-seg \cite{opensistemas_yolov8_crack_seg}, and SCSegamba \cite{liu2025scsegamba} for crack. 
Crucially, the platform provides interactive tools, such as point-based prompting and brush-based refinement, allowing human experts to perform fine-grained correction on model-generated masks. 
This hybrid approach ensures that even amorphous defects, such as irregular staining or peeling, possess pixel-level geometric accuracy. 
The interface and operational workflow of our interactive platform are detailed in Fig.\ref{figs1:ui} of the Appendix.

Through this rigorous curation and annotation pipeline, we establish DefectBench, a comprehensive dataset specifically tailored for building facade inspection. 
To the best of our knowledge, DefectBench stands as the most comprehensive and granular resource in this domain, uniquely providing synchronized, pixel-level ground truth across the full spectrum of classification, object detection, and semantic segmentation.
The finalized dataset comprises 1,488 annotated images, encompassing a diverse array of 4,527 distinct structural anomalies.
The detailed distribution of defect instances and statistical description are summarized in Table \ref{tab1:defect_statistics}.

\begin{table}[t]
\centering
\caption{Statistical distribution of facade defect categories}
\label{tab1:defect_statistics}
\begin{tabular}{ccc}
\hline
Primary   class                   & Sub-type             & Number \\ \hline
\multirow{2}{*}{Crack}            & Linear crack         & 2042   \\
                                  & Map cracking         & 80     \\
\multirow{2}{*}{Material   loss}  & Spalling             & 906    \\
                                  & Peeling              & 51     \\
\multirow{3}{*}{Surface stain}    & Corrosion            & 254    \\
                                  & Rust stain           & 21     \\
                                  & Leakage stain        & 803    \\
\multirow{3}{*}{External fixings} & Vegetation   growth  & 221    \\
                                  & Graffiti             & 30     \\
                                  & Surface contaminants & 119    \\ \hline
\end{tabular}
\end{table}

\subsection{Hierarchical Benchmarking Tasks}
Based on the established dataset, we formulate the building facade inspection task as a hierarchical multimodal understanding problem. 
Given a facade image $I$ and a task-oriented language prompt $P$, the objective is to leverage a LMM to generate a structured response $R$ that encompasses semantic, spatial, and geometric dimensions.

\paragraph{Semantic Perception: The ``What'' Task}
The first level of diagnosis focuses on identifying the nature and quantity of anomalies. 
We define the semantic perception task as:
\[
R_{\text{sem}} = \mathcal{F}_{\text{LMM}}(I, P_{\text{cls}})
\]
where $P_{\text{cls}}$ queries the model for defect classification (e.g., crack, material loss, surface stain, or human damage) and defect counting. 
Unlike traditional closed-set classification, this task requires the model to map visual features to expert-level engineering terminology under a zero-shot or few-shot setting.

To enable actionable maintenance, the model must resolve the spatial layout and structural context of defects on the facade. 
We formulate spatial localization as a dual-component task that bridges geometric coordinates with semantic topology. 
Specifically, the model is required to generate precise bounding boxes 
$B = \{x_1, y_1, x_2, y_2\}$ for the identified anomalies, while simultaneously performing visual spatial reasoning to parse the topological relationships among defects, such as cracks radiating from a spalling area.

The overall mapping is defined as:
\[
R_{\text{spa}} = \mathcal{F}_{\text{LMM}}(I, P_{\text{loc}})
\]
where $R_{\text{spa}}$ comprises both spatial coordinates and relational descriptions.

\paragraph{Geometry Segmentation: The ``How'' Task}
The most fine-grained level of the benchmark focuses on the precise morphology of defects. 
We define the geometry segmentation task as the generation of a binary pixel-level mask
$M \in \{0,1\}^{H \times W}$.
Given an input image--prompt pair $(I, P_{\text{seg}})$, the model is required to perform dense prediction:
\[
M_{\text{Seg}} = \mathcal{F}_{\text{LMM}}(I, P_{\text{seg}})
\]
where the ``How'' task evaluates the model’s ability to delineate irregular and fine-grained topologies, such as the amorphous boundaries of surface staining or the jagged edges of material loss, which are substantially more complex than standard man-made objects.

\section{Benchmark and Experiments}

\subsection{Experimental Settings}
\subsubsection{Baseline Models}
o provide a comprehensive assessment of LMMs in building facade inspection, we evaluate a diverse suite of 24 representative models. This includes 12 closed-source models (e.g., Claude-Opus-4.1 \cite{anthropic_claude4_model_card}, Gemini-3-Pro-Preview \cite{deepmind_gemini3pro_model_card}, Gemini-3-Flash-Preview, GPT-5.2-Pro, GPT-5.2-Chat-Latest \cite{openai_5_2_system_card}, GPT-4o \cite{achiam2023gpt}, and Doubao-Seed-1.8 \cite{seed1_8_modelcard}) and 12 open-source models (e.g., Qwen2.5/3-VL series \cite{qwen25_vl_collection,qwen3_vl_collection}, GLM series\cite{zeng2025glm,zai_glm46_model_card}, LLaVa-34B\cite{liu2023visual}, InternVL3.5\cite{wang2025internvl3}, DeepSeek-VL2\cite{wu2024deepseek}, and Kimi-K2.5\cite{kimiteam2026kimik25visualagentic}). 
These models are selected to reflect the current state-of-the-art in multimodal understanding and reasoning.

Since \textit{DefectBench} assesses LMMs across three hierarchical dimensions of varying complexity, we employ task-specific metrics for each level.
For Semantic Perception (Level 1), we adopt Precision (P), Recall (R), and F1-score (F1.) for category identification, while utilizing Mean Absolute Error (MAE) and Relative Error (RE) to quantify the accuracy of numerical defect counting.
Transitioning to Spatial Localization (Level 2), the evaluation shifts toward visual grounding and topological reasoning. We employ mAP$_{50}$, mAP$_{50-95}$, and F1. to measure the precision of generated bounding boxes. Furthermore, the performance of visual spatial reasoning is quantified through P., R., and HR, specifically evaluating the logical correctness of the predicted relational descriptions between structural components.
In terms of Geometry Segmentation (Level 3), we evaluate the morphological fidelity of the generated binary masks.
Thus we report the mean Intersection over Union (mIoU), P., R., F1., and Pixel Accuracy (PA). 
These multi-tiered metrics ensure a rigorous and comprehensive quantification of LMM performance, ranging from coarse-grained semantics to fine-grained geometric reconstruction.


\subsubsection{Model Deployment}

To ensure a fair evaluation across diverse defect morphologies, we sampled randomly approximately 100 images to maintain a near-uniform distribution from each primary defect category to maintain a near-uniform distribution, thereby mitigating potential evaluation bias toward dominant classes. 
The resulting evaluation set comprises 487 high-quality samples, with a balanced categorical breakdown: Crack (117), Material Loss (102), Surface Stain (134), and External Fixing (122).
The benchmarking process was executed through a dual-modality deployment framework. Closed-source models and specific high-parameter open-source models were accessed via remote cloud service APIs to ensure deterministic outputs. In parallel, other open-source models were deployed on a high-performance local server equipped with four NVIDIA RTX 4090 GPUs. All local experiments were implemented using the PyTorch framework.

\subsection{Results and Analysis}

\subsubsection{Semantic Understanding (What)}
The experimental results for the Level 1 task, summarized in Table \ref{tab2:level1}, reveal several critical insights into the zero-shot capabilities of LMMs in the structural engineering domain.
A pivotal finding is that while closed-source models have historically led in general reasoning, current-generation open-source LMMs, notably the Qwen3, GLM, and InternVL series, have begun to demonstrate superior stability and numerical consistency in deterministic detection and identification tasks.
While the Gemini-3 series achieves impressive classification metrics, particularly in terms of recall ($R=0.9711$) and hit rate ($HR=0.9938$), \textit{Qwen3-VL-32B-Thinking} and \textit{GLM-4.6V} deliver the most competitive performance. 
The analysis of the Qwen series confirms that performance gains generally align with the Scaling Law from 7B to 72B.
Notably, Qwen3-VL-32B-Thinking—equipped with an enhanced reasoning chain—surpasses the 72B variant in $HR$ ($0.9840$). 
Furthermore, the lightweight Qwen3-VL-8B performs on par with Qwen2.5-VL-72B. 
These findings suggest that for specialized diagnostic tasks in complex structural environments, the integration of logical reasoning mechanisms is more efficacious than a mere increase in parameter scale.

Additionally, while most models achieve a baseline $HR$ above $0.6$ for defect identification (Q1), their performance fluctuates drastically in the numerical counting task (Q2). 
Excluding Claude, most models (including GPT-4o and Gemini-3 Pro) exhibit an MAE near or exceeding $1.0$. 
This phenomenon indicates that when faced with multiple symbiotic micro-defects, LMMs frequently suffer from redundant counts or omissions. 
Although the overall deviation remains within one unit of error, these results highlight a fundamental limitation in the models' current capacity for precise spatial quantification and logical enumeration.
Also, the models exhibit polarized performance across different defect taxonomies. Crack and External Fixings represent the most bifurcated categories; while Claude achieves near-perfect MAE ($0.0588$ and $0.0444$), lightweight models like Qwen2.5-VL-7B show high Relative Error (RE). 
Conversely, the Surface Stain category poses a universal challenge, yielding consistently high MAE across all tested models. 
This difficulty is attributed to the inherent visual ambiguity of stains, characterized by blurred boundaries and irregular topologies, which remains a formidable "semantic hurdle" even for LMMs.

\begin{table*}[t]
\centering
\resizebox{\textwidth}{!}{
\begin{tabular}{cccccccccccccc}
\hline
\multirow{3}{*}{Models} & \multicolumn{3}{c}{\multirow{2}{*}{Q1: What defects are in the   image?}} & \multicolumn{10}{c}{Q2: How many instances of each defect type?}                                                                                                                                                                                                                                                                        \\
                        & \multicolumn{3}{c}{}                                                      & \multicolumn{2}{c}{Crack}                                       & \multicolumn{2}{c}{Material loss}                               & \multicolumn{2}{c}{Surface stain}                               & \multicolumn{2}{c}{External fixings}                            & \multicolumn{2}{c}{Average}                                     \\
                        & P.                              & R.                             & F1.    & MAE                            & RE                             & MAE                            & RE                             & MAE                            & RE                             & MAE                            & RE                             & MAE                            & RE                             \\ \hline
\multicolumn{14}{c}{\textit{\textbf{Open-Source Large Multimodal Model}}}                                                                                                                                                                                                                                                                                                                                                                     \\ \hline
GLM-4.5V                & 0.6698                          & 0.8264                         & 0.7399 & 0.7454                         & 0.4194                         & 0.4559                         & \textbf{0.2435}                & 0.9815                         & 0.5342                         & 0.3285                         & 0.3986                         & 0.6278                         & 0.3989                         \\
GLM-4.6V                & 0.7249                          & 0.7778                         & 0.7504 & 0.7290                         & 0.4405                         & \textbf{0.4435}                & 0.2774                         & 1.0103                         & 0.6125                         & \textbf{0.2053}                & 0.3477                         & \textbf{0.5970}                & 0.4195                         \\
Deepseek-vl2            & 0.6191                          & 0.5174                         & 0.5637 & 0.9322                         & 0.4115                         & 0.5544                         & 0.3885                         & 1.1047                         & 0.8788                         & 0.3778                         & 0.8294                         & 0.7423                         & 0.6271                         \\
InternVL3.5             & 0.5646                          & 0.6528                         & 0.6055 & 0.8522                         & 0.4169                         & 0.5934                         & 0.3041                         & 1.2361                         & 0.7037                         & 0.5195                         & 0.8033                         & 0.8003                         & 0.5570                         \\
LLaVa-34B               & 0.6068                          & 0.5984                         & 0.6026 & 0.6838                         & 0.4771                         & 0.7906                         & 0.4913                         & 1.0698                         & 0.7419                         & 0.3244                         & 0.4944                         & 0.7172                         & 0.5512                         \\
Qwen2.5-VL-72B-instruct & 0.6912                          & 0.7720                         & 0.7294 & 0.8563                         & 0.5165                         & 0.5154                         & 0.3161                         & 1.0287                         & 0.6086                         & 0.2608                         & 0.4869                         & 0.6653                         & 0.4820                         \\
Qwen2.5-VL-7B-instruct  & 0.5240                          & 0.4421                         & 0.4796 & 0.9856                         & 0.6714                         & 0.8419                         & 0.6271                         & 1.1027                         & 0.6351                         & 0.4969                         & 0.9125                         & 0.8568                         & 0.7115                         \\
Qwen2.5-VL-32B-instruct & 0.5948                          & 0.7627                         & 0.6684 & 0.9138                         & 0.3602                         & 0.6386                         & 0.3377                         & \textbf{0.9425}                & 0.5883                         & 0.3018                         & 0.4716                         & 0.6992                         & 0.4395                         \\
Qwen3-VL-8B-instruct    & 0.6784                          & 0.7812                         & 0.7262 & \textbf{0.6160}                & 0.4828                         & 0.5318                         & 0.2457                         & 0.9528                         & 0.5033                         & 0.3347                         & 0.4744                         & 0.6088                         & 0.4266                         \\
Qwen3-VL-32B-thinking   & \textbf{0.7431}                 & \textbf{0.9772}                & \textbf{0.8442} & 1.2019                         & 0.5689                         & 0.4808                         & 0.3282                         & 1.0865                         & 0.7682                         & 0.2692                         & \textbf{0.2188}                & 0.7596                         & 0.4710                         \\
Qwen3-VL-32B-instruct   & 0.5748                          & 0.9514                         & 0.7166 & 0.9261                         & \textbf{0.3506}                & 0.8398                         & 0.4826                         & 1.1499                         & \textbf{0.4198}                & 0.3593                         & 0.3316                         & 0.8188                         & \textbf{0.3962}                \\ \hline
\multicolumn{14}{c}{\textit{\textbf{Closed-Source Large Multimodal Model}}}                                                                                                                                                                                                                                                                                                                                                                   \\ \hline
gpt-4o                  & 0.6983                          & 0.8252                         & 0.7565 & 1.1569                         & 0.5207                         & 0.7179                         & 0.4226                         & 1.3386                         & 0.5500                         & 0.6611                         & 0.5386                         & 0.9686                         & 0.5080                         \\
gpt-5.1-chat-latest     & {\ul \textit{\textbf{0.7267}}}  & 0.9016                         & \textbf{0.8048} & 1.2418                         & 0.5271                         & 0.5429                         & 0.3008                         & 1.1076                         & {\ul \textit{\textbf{0.4038}}} & 0.5556                         & 0.4200                         & 0.8620                         & 0.4129                         \\
gpt-5.2-pro             & 0.6349                          & 0.9097                         & 0.7479 & 1.0392                         & 0.4909                         & {\ul \textit{\textbf{0.5071}}} & {\ul \textit{\textbf{0.2827}}} & 1.2151                         & 0.5091                         & 0.7111                         & 0.5960                         & 0.8495                         & 0.4506                         \\
gemini-3-flash-preview  & 0.5665                          & {\ul \textit{\textbf{0.9711}}} & 0.7156 & 1.1634                         & 0.5763                         & 0.7286                         & 0.4995                         & 1.2351                         & 0.6551                         & 1.0889                         & 0.9812                         & 1.0540                         & 0.6780                         \\
gemini-3-pro-preview    & 0.6158                          & 0.9294                         & 0.7408 & {\ul \textit{\textbf{1.0196}}} & {\ul \textit{\textbf{0.4393}}} & 0.5964                         & 0.3796                         & {\ul \textit{\textbf{1.0637}}} & 0.4735                         & 0.7778                         & 0.6852                         & {\ul \textit{\textbf{0.8449}}} & {\ul \textit{\textbf{0.4811}}} \\
Doubao-Seed-1.8         & 0.6276                          & 0.8935                         & 0.7373 & 1.6405                         & 0.6990                         & 0.6321                         & 0.2936                         & 1.4542                         & 0.5511                         & {\ul \textit{\textbf{0.5000}}} & {\ul \textit{\textbf{0.3869}}} & 1.0567                         & 0.4827                         \\
Claude-opus-4.1         & 0.6809                          & 0.8843                         & 0.7694 & 1.1176                         & 0.6270                         & 0.7714                         & 0.5106                         & 1.2191                         & 0.5062                         & 0.6556                         & 0.5205                         & 0.9387                         & 0.5320                         \\ \hline
\end{tabular}
}
\caption{Performance benchmark of 16 state-of-the-art LMMs on the Level 1 (Semantic Understanding) task of DefectBench. Evaluation covers defect classification accuracy (Q1) and numerical counting precision (Q2) across four primary categories.}
\label{tab2:level1}
\end{table*}

\subsubsection{Spatial Localization (Where)}
The experimental results for the Level 2 task detailed in Table \ref{tab3:q3_q4_results}, evaluate the models' proficiency in spatial grounding and topological reasoning for building defect inspection.
The results reveal that Accurate Object Detection (Q3) remains a significant challenge for contemporary LMMs in a zero-shot setting.
Although Gemini-3-pro-preview achieves the highest overall performance with a lead in Recall ($R = 0.5333$) and F1-score ($F1 = 0.4226$), there is a significant performance gap persists when compared to task-specific, supervised expert models \cite{benz2022image,zha2025dataset}.
Notably, models specifically architected for enhanced logical reasoning (such as Qwen3-VL-32B-Thinking, the GPT series, and Claude-Opus-4.1) demonstrate a performance collapse in coordinate-based localization.

In contrast to the geometric struggles in Q3, models demonstrate a much stronger grasp of Visual Spatial Reasoning (Q4).
The performance divergence between detection tasks (Q3) and visual spatial reasoning (Q4) highlights a fundamental characteristic of current LMMs: while they lack the geometric fidelity required for fine-grained coordinate localization, they exhibit robust topological awareness regarding structural anomalies.
First, closed-source models consistently lead in contextualizing the spatial distribution of facade deterioration.
Gemini-3-pro-preview establishes the performance ceiling with an F1-score of $0.8390$, effectively parsing complex structural dependencies and relational hierarchies. 
However, a systematic precision-recall asymmetry is observed across the benchmark.
While models maintain high fidelity in characterizing detected topological relationships, they suffer from a significant false negative rate. 
This implies that despite the models effectively decode the relationship of local defects, they struggle to maintain global scene consistency.
In the open-source domain, InternVL3.5 achieves the benchmark's highest precision ($P = 0.8462$) and lower recall ($R = 0.2112$).
Conversely, GLM-4.6V demonstrates robust versatility, balancing relational accuracy with broader structural perception. 
These findings corroborate that contemporary LMM architectures are inherently better suited for qualitative diagnostic inference rather than quantitative spatial mapping.



\begin{table}[t]
\centering
\resizebox{\columnwidth}{!}{
\begin{tabular}{cccccll}
\hline
\multirow{2}{*}{Model}  & \multicolumn{3}{c}{Q3:   Object Detection}                                                       & \multicolumn{3}{c}{Q4:   Visual Spatial Reasoning}                                                                                            \\
                        & P.                             & R.                             & F1.                            & P.                                  & \multicolumn{1}{c}{R.}                             & \multicolumn{1}{c}{F1.}                            \\ \hline
\multicolumn{7}{c}{\textit{\textbf{Open-Source Large   Multimodal Model}}}                                                                                                                                                                                                 \\ \hline
GLM-4.5V                & 0.3157                         & 0.3866                         & 0.3476                         & \multicolumn{1}{l}{0.7131}          & 0.6198                                             & 0.6632                                             \\
GLM-4.6V                & 0.3587                         & \textbf{0.3981}                & \textbf{0.3774}                & \multicolumn{1}{l}{0.7093}          & \textbf{0.6653}                                    & \textbf{0.6866}                                    \\
LLaVa-34B               & 0.2871                         & 0.1366                         & 0.1851                         & \multicolumn{1}{l}{0.2246}          & 0.1353                                             & 0.1688                                             \\
InternVL3.5             & 0.2517                         & 0.1725                         & 0.2047                         & \multicolumn{1}{l}{\textbf{0.8462}} & 0.2112                                             & 0.3380                                             \\
Qwen2.5-VL-7B-Instruct  & 0.2949                         & 0.0447                         & 0.0777                         & \multicolumn{1}{l}{0.2630}          & 0.1194                                             & 0.1642                                             \\
Qwen2.5-VL-32B-Instruct & 0.3385                         & 0.2706                         & 0.3008                         & \multicolumn{1}{l}{0.4462}          & 0.3172                                             & 0.3708                                             \\
Qwen2.5-VL-72B-Instruct & \textbf{0.3834}                & 0.2429                         & 0.2974                         & \multicolumn{1}{l}{0.6815}          & 0.5666                                             & 0.6188                                             \\
Qwen3-VL-8B-Instruct    & 0.3062                         & 0.3029                         & 0.3045                         & \multicolumn{1}{l}{0.4460}          & 0.2094                                             & 0.2850                                             \\
Qwen3-VL-32B-Instruct   & 0.2227                         & 0.2464                         & 0.2340                         & \multicolumn{1}{l}{0.5879}          & 0.5325                                             & 0.5588                                             \\
Qwen3-VL-32B-Thinking   & 0.1457                         & 0.0417                         & 0.0648                         & 0.4345                                   & \multicolumn{1}{c}{0.5652}                              & 0.4763                              \\ \hline
\multicolumn{7}{c}{\textit{\textbf{Closed-Source Large   Multimodal Model}}}                                                                                                                                                                                               \\ \hline
gpt-4o                  & 0.0098                         & 0.0114                         & 0.0104                         & 0.6097                              & \multicolumn{1}{c}{0.5698}                         & \multicolumn{1}{c}{0.5891}                         \\
gpt-5.1-chat-latest     & 0.1027                         & 0.1056                         & 0.1039                         & 0.7314                              & \multicolumn{1}{c}{0.5846}                         & \multicolumn{1}{c}{0.6499}                         \\
gpt-5.2-pro             & 0.2089                         & 0.2870                         & 0.2357                         & 0.7989                              & \multicolumn{1}{c}{0.7941}                         & \multicolumn{1}{c}{0.7965}                         \\
Gemini-3-flash-preview  & 0.2380                         & 0.4974                         & 0.3188                         & 0.8366                              & \multicolumn{1}{c}{0.5899}                         & \multicolumn{1}{c}{0.6919}                         \\
Gemini-3-pro-preview    & {\ul \textit{\textbf{0.3607}}} & {\ul \textit{\textbf{0.5333}}} & {\ul \textit{\textbf{0.4226}}} & {\ul \textit{\textbf{0.8655}}}      & \multicolumn{1}{c}{{\ul \textit{\textbf{0.8141}}}} & \multicolumn{1}{c}{{\ul \textit{\textbf{0.8390}}}} \\
Doubao-Seed-1.8         & 0.3487                         & 0.4702                         & 0.3969                         & 0.6305                              & \multicolumn{1}{c}{0.6684}                         & \multicolumn{1}{c}{0.6489}                         \\
Claude-opus-4.1         & 0.1608                         & 0.2243                         & 0.1867                         & 0.7615                              & \multicolumn{1}{c}{0.7801}                         & \multicolumn{1}{c}{0.7707}                         \\ \hline
\end{tabular}
}
\caption{Performance comparison on object detection and visual spatial reasoning tasks.}
\label{tab3:q3_q4_results}
\end{table}

\subsubsection{Generative Geometry Segmentation (How)}
Table \ref{tab4: seg} presents the quantitative evaluation of pixel-level segmentation. 
Crucially, this task represents a paradigm shift from traditional discriminative feature extraction to \textbf{generative segmentation}, where models synthesize precise object masks conditioned on multimodal prompts. 
Unlike the semantic reasoning tasks discussed previously, this paradigm demands a dual competency: the semantic understanding to interpret "what" to segment, and the geometric precision to delineate "where" the boundaries lie.

The evaluated SOTA LMMs demonstrate remarkable proficiency in this challenging domain. 
Gemini-3-pro-edit, powered by the Nano Banana Pro Edit architecture, establishes the empirical upper bound for the benchmark. 
It achieves a dominant mIoU of $\mathbf{0.7418}$ and an F1-score of $\mathbf{0.7875}$, with balanced Precision ($0.8467$) and Recall ($0.8072$). 
This substantial margin suggests that Gemini possesses a superior architectural bias for semantic-geometric alignment, effectively preserving fine-grained boundary details while maintaining global consistency. 
Also, the remaining SOTA LMMs exhibit a convergent performance trend, with mIoU scores clustering in the $0.68 - 0.69$ range in a completely zero-shot setting \cite{liu2025scsegamba,khanhha_crack_seg_dataset}.
Their performance is competitive with specialized supervised segmentation models. 
These findings underscore the transformative potential of LMMs in automated defect inspection, offering precise localization without the need for task-specific training data.


\begin{table}[t]
\resizebox{\columnwidth}{!}{
\begin{tabular}{cccccc}
\hline
Model                & mIoU            & P.              & R.              & F1.             & PA              \\ \hline
Qwen-image-edit-plus & 0.6852          & 0.8110          & 0.7495          & 0.7265          & 0.9709          \\
Wan2.6-image         & 0.6848          & 0.8121          & 0.7349          & 0.7192          & \textbf{0.9717} \\
Doubao-seedream-4.5  & 0.6920          & 0.7867          & 0.7442          & 0.7344          & 0.9676          \\
Gemini-3-pro-edit    & \textbf{0.7418} & \textbf{0.8467} & \textbf{0.8072} & \textbf{0.7875} & 0.9713          \\
FLUX-2-pro-edit      & 0.6926          & 0.8032          & 0.7455          & 0.7397          & 0.9693          \\ \hline
\end{tabular}
}
\caption{Quantitative comparison of pixel-level segmentation performance across SOTA LMMs.}
\label{tab4: seg}
\end{table}

\subsubsection{Error Analysis and Limitation}


To emulate authentic diagnostic workflows, our benchmark adopts a multi-turn dialogue protocol where subsequent queries are explicitly conditioned on the model's preceding outputs (Prompt shown in the appendix Table \ref{tab:prompts}).
While this design preserves contextual continuity, it inherently introduces the risk of cascading error propagation. 
Specifically, hallucinations or misclassifications emerging during the Global Perception phase (Q1, Q2) establish erroneous semantic priors, which are subsequently integrated into the prompting context for Object Detection (Q3).
As empirically observed in the evaluation of Qwen3-32B-Instruct (illustrated in Appendix Fig. \ref{figs2:det}), this context-induced bias frequently results in "right location, wrong class" anomalies, where bounding boxes are spatially accurate but semantically misaligned.

\begin{figure}[ht]
  \centering
  \includegraphics[width=1\linewidth]{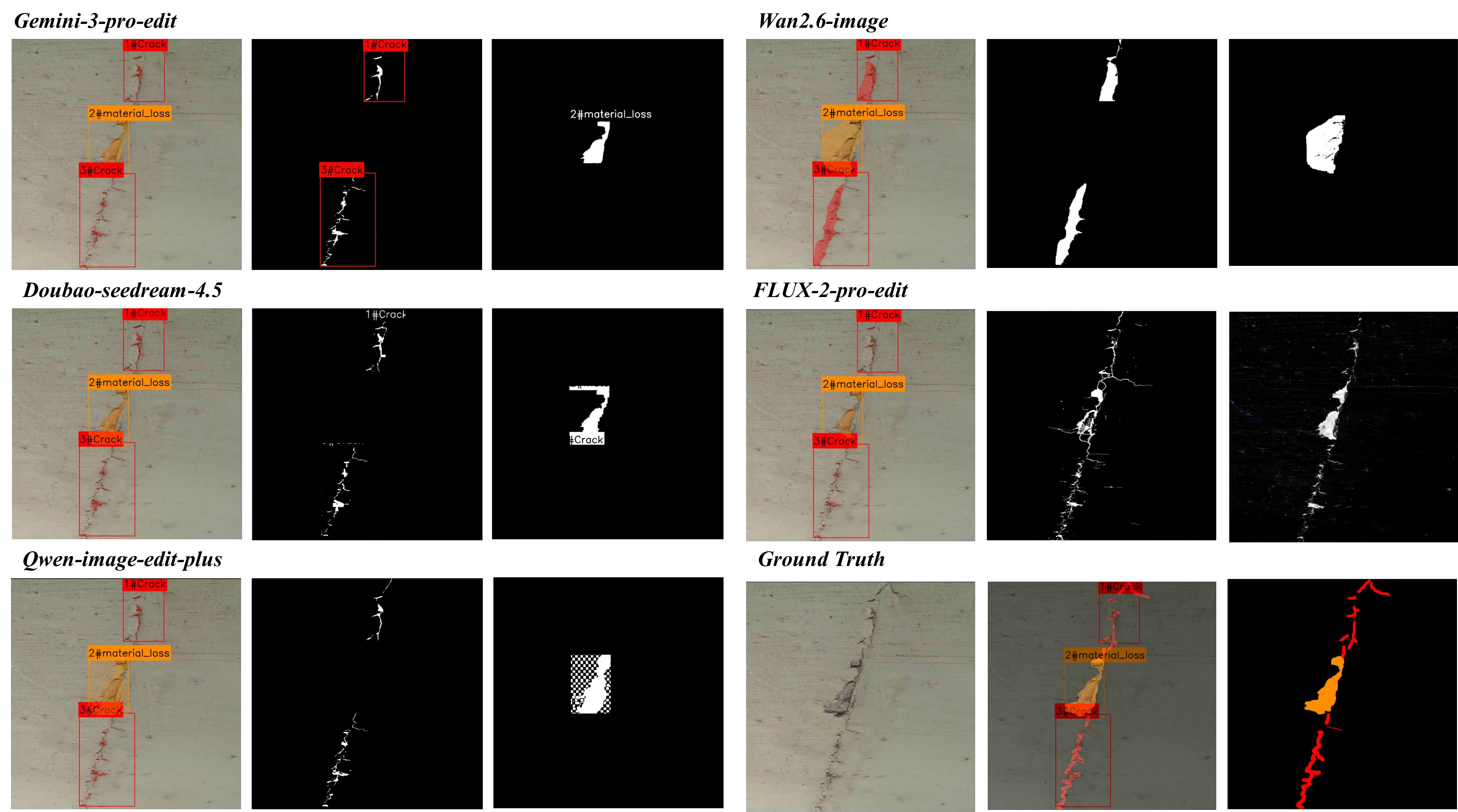}
  \caption{Visualization of generative instability and instruction alignment failures in zero-shot segmentation.}
  \label{fig2:det}
\end{figure}

\begin{figure}[ht]
  \centering
  \includegraphics[width=1\linewidth]{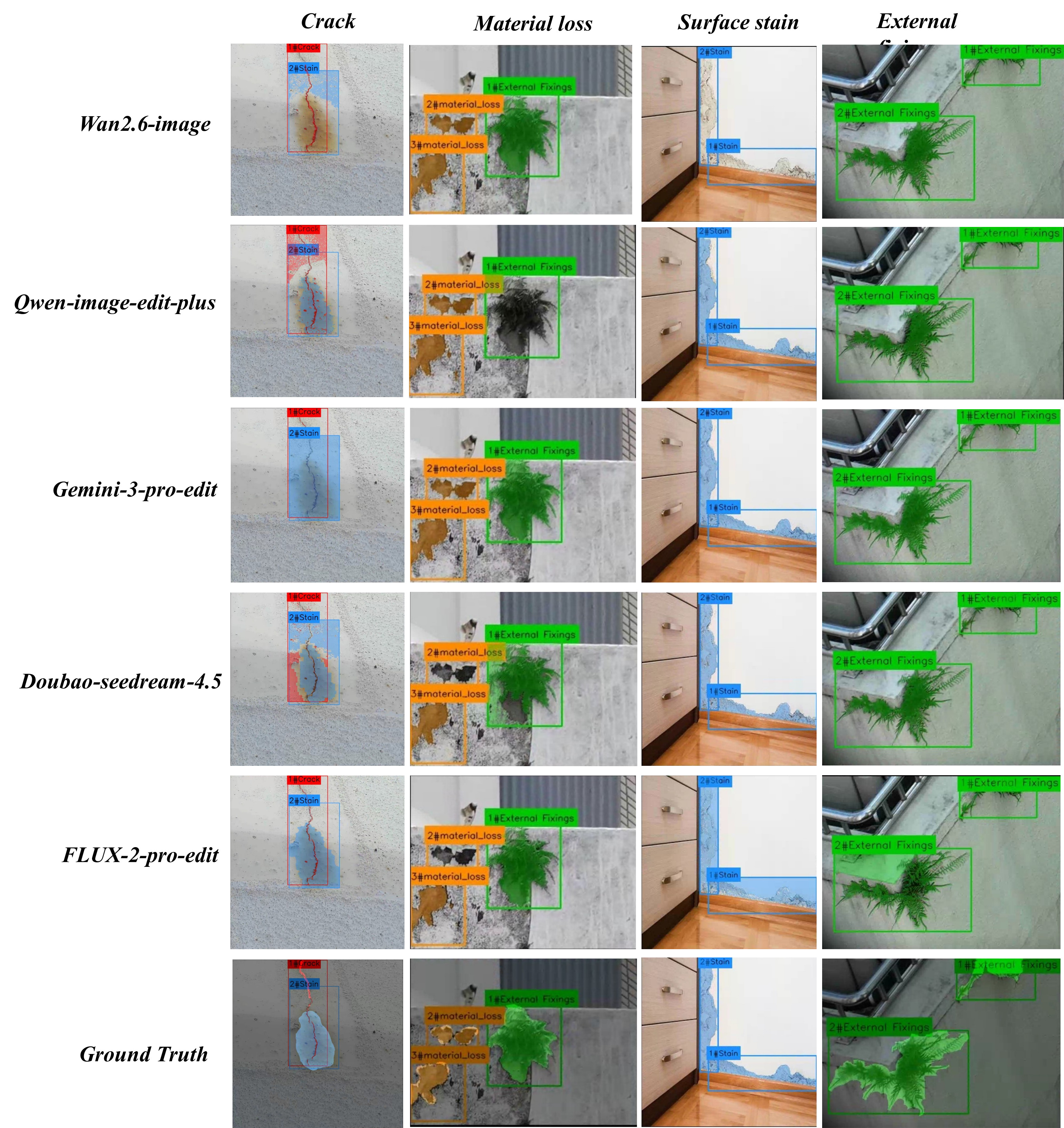}
  \caption{Cross-model qualitative evaluation of generative geometry segmentation across multiple facade defect categories.}
  \label{fig3:seg}
\end{figure}

The third tier of our benchmark utilizes visual prompting mechanisms to spatially constrain generative geometry segmentation.
While quantitative results assessments (Table \ref{tab4: seg}) reveal promising topological awareness, current LMMs are frequently plagued by inherent generative stochasticity and instruction alignment failures. 
For instance, despite explicit system directives for binary mask synthesis, models often exhibit spectral hallucinations, erroneously generating RGB textures or reconstructing irrelevant background context instead of producing clean segmentation maps (illustrated in Fig.\ref{fig2:det}).
Therefore, we integrated a deterministic post-processing pipeline to rigorously address these generative anomalies.
This module enforces adaptive binarization on the raw outputs and spatially gates the evaluation to the Region of Interest (RoI) established by the detection priors.
This methodological constraint effectively decouples geometric fidelity from generative noise, ensuring that the evaluation metrics exclusively quantify the model's capacity for precise boundary delineation and topological understanding, independent of background artifacts.
As illustrated in Fig. \ref{fig3:seg}, the generative capabilities of LMMs for geometry segmentation reveal a compelling paradox. 
On one hand, models demonstrate exceptional sub-pixel boundary adherence, often capturing irregular topological nuances with a precision that surpasses human-annotated ground truth. 
This highlights the immense potential of LMMs for automated, high-fidelity defect mapping. 
On the other hand, hallucination-induced artifacts persist, manifesting as occasional omissions of subtle features or false positives in complex texture regions, underscoring the need for further robustness in constrained generation tasks.

\section{Conclusion}
In this work, we introduced DefectBench, a comprehensive framework evaluating LMMs for facade defect inspection across three hierarchical levels: semantic perception, spatial localization, and geometric segmentation. 
Our extensive experiments on 18 LMMs reveal a fundamental dichotomy between semantic reasoning and metric precision. While state-of-the-art models (e.g., Gemini-3) demonstrate exceptional topological awareness, they struggle with precise spatial grounding, suggesting their current utility lies more in qualitative inference than quantitative mapping.
Crucially, we validate the viability of zero-shot generative segmentation, showing that foundation models can rival specialized supervised networks without domain-specific training.
Besides, we release a unified multi-task dataset meticulously annotated for classification, detection, and segmentation, which serves as a holistic testbed for assessing the multifaceted zero-shot capabilities of LMMs. 
Although challenges such as cascading error propagation and generative hallucinations persist, our study establish a new baseline for automated infrastructure monitoring.

\bibliographystyle{acm}
\bibliography{software}

\appendix

\section{Appendix}

\begin{figure}[H]
  \centering
  \includegraphics[width=0.9\linewidth]{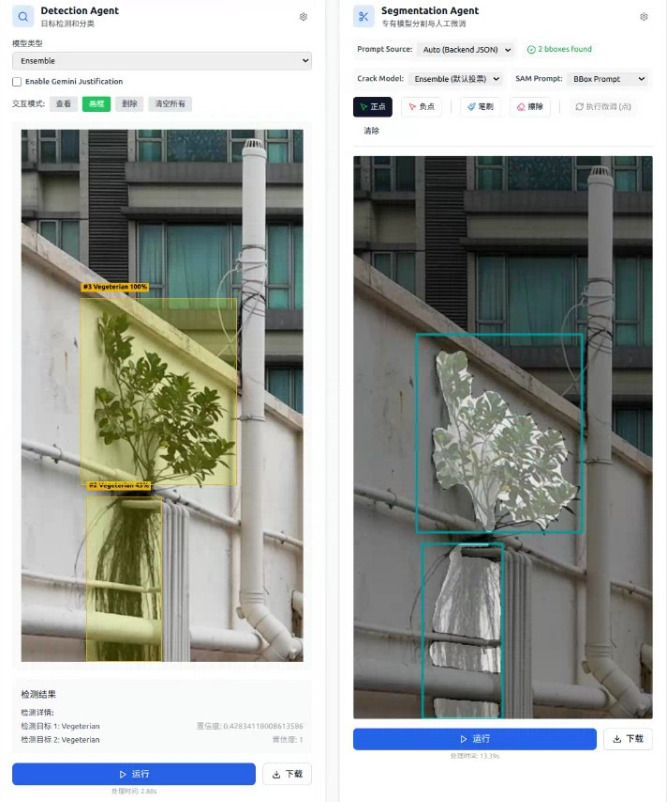}
  \caption{UI for plug-and-play annotation platform.}
  \label{figs1:ui}
\end{figure}

\begin{figure}[H]
  \centering
  \includegraphics[width=0.8\linewidth]{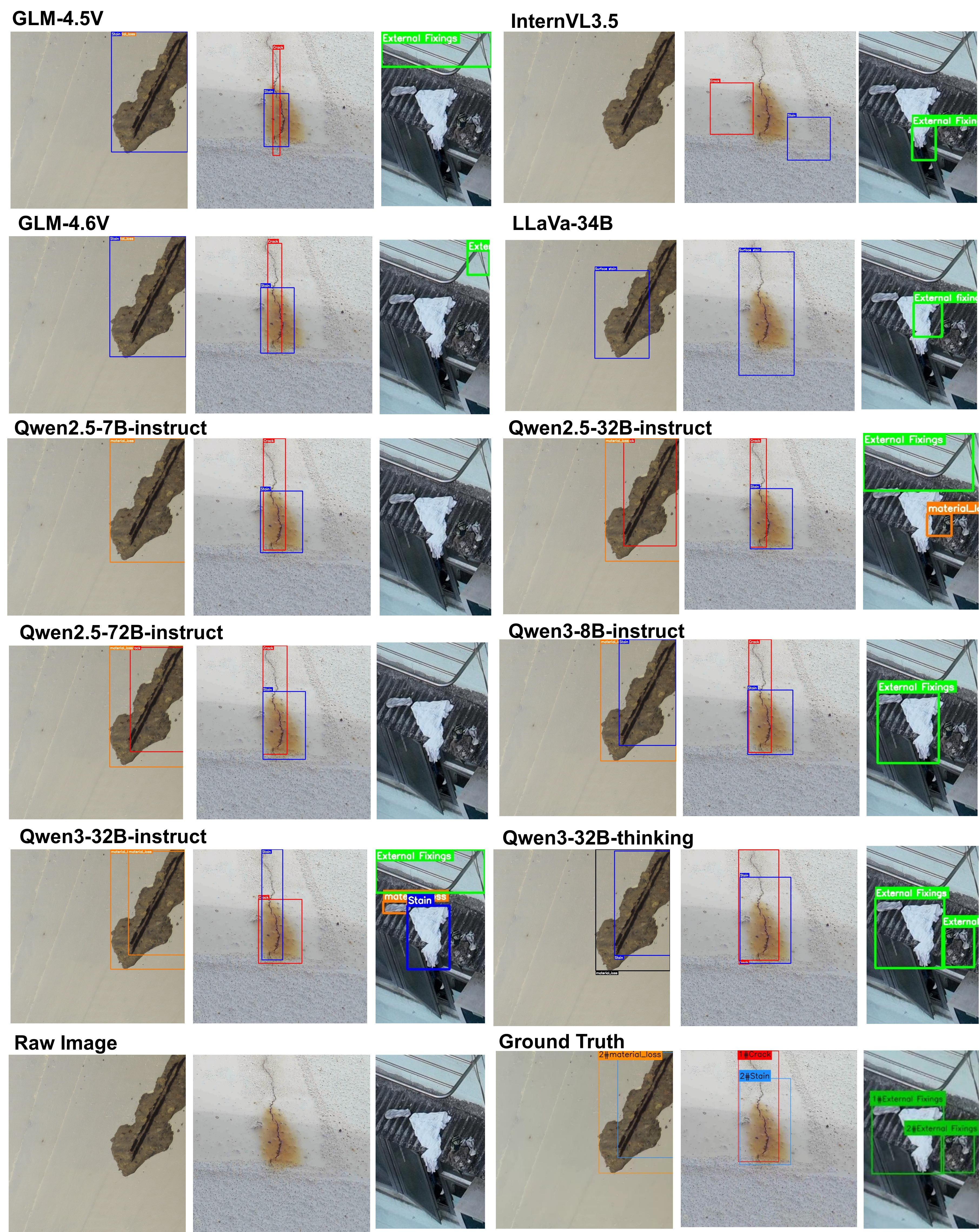}
  \caption{Comparative defect detection results of different LMMs on facade images.}
  \label{figs2:det}
\end{figure}

\begin{table*}[t]
\centering
\resizebox{\textwidth}{!}{
\begin{tabular}{lllll}
\hline
Task                            & Dataset   & Number & Category                                                                                                                                  & Annotation Level   \\ \hline
\multirow{4}{*}{classification} & CCIC      & 20000  & surface cracks                                                                                                                            & Patch-level        \\
                                & BDD       & 436    & Three categories: roof defect,   cracks defect and flakes                                                                                 & Patch-level        \\
                                & BD3       & 3965   & Six categories: algae, major   crack, minor crack, peeling, spalling, and stain                                                           & Patch-level        \\
                                & HS-23K    & 23,688 & Six categories: biological   deterioration, chemical deterioration, cracks, human-induced damage, material   loss, and undamaged surfaces & Patch-level        \\
\multirow{3}{*}{Detection}      & MBDD2025  & 14,471 & Five categories: crack, leakage,   abscission, corrosion, bulge                                                                           & Bounding-box level \\
                                & BDW       & 1417   & Five categories: crack, mold,   peeling, stairstep crack, water seepage                                                                   & Bounding-box level \\
                                & CUBIT-Det & 5,527  & Three categories: crack,   spalling, moisture                                                                                             & Bounding-box level \\
\multirow{5}{*}{segmentation}   & Bai-2020  & 1221   & Crack                                                                                                                                     & Pixel-level        \\
                                & CSD       & 11,298 & Crack                                                                                                                                     & Pixel-level        \\
                                & DeepCrack & 537    & Crack                                                                                                                                     & Pixel-level        \\
                                & CUBIT-Seg & 6622   & Crack, spalling                                                                                                                           & Pixel-level        \\
                                & S2DS      & 743    & Five categories: crack, spalling,   corrosion, efflorescence, vegetation                                                                  & Pixel-level        \\ \hline
\end{tabular}
}
\caption{Statistics of collected open-source datasets for building defects}
\label{tab:s1}
\end{table*}

\begin{table*}[t]
\centering
\renewcommand{\arraystretch}{1.2}
\small
\begin{tabular}{p{0.12\textwidth} p{0.83\textwidth}}
\hline
\textbf{Question} & \textbf{Prompt} \\
\hline

Q1: Defect Identification &
You are an expert in building defect analysis. You will see one image.
Your task is to answer two questions about visible defects in this image.

Only consider the following four primary defect types, and use \textbf{exactly} these English names:
Crack; material\_loss; Stain; External Fixings.

Return your answers strictly as a single JSON object:
\texttt{\{"answer1": "...", "answer2": "..."\}}.
The JSON must be valid and parseable by \texttt{json.loads}. \\

\hline

Q2: Defect Counting &
Based on the identified defect types, report the number of instances for each defect class.
Counts must follow the same order as in Q1 and be separated by commas. \\

\hline

Q3: Object Detection &
Detect bounding boxes for each defect instance identified previously.
Each object must be returned with its category and bounding box in the format:
\texttt{bbox: "<x\_min y\_min x\_max y\_max>"}.

Return \textbf{only} a JSON array:
\texttt{[{"category": "...", "bbox": "..."}]}.
Use exactly the category names: Crack, material\_loss, Stain, External Fixings. \\

\hline

Q4: Visual Spatial Reasoning &
Analyze spatial relationships between all pairs of detected defects.
Possible relations include: inclusion, overlapping, adjacency, and disjoint.

Return a JSON object with the structure:
\texttt{\{"relationships": [[i\#type, relation, j\#type]]\}}.
The output must be valid JSON without any additional text. \\

\hline

Q5: Geometry Segmentation &
Edit the input image directly to generate a binary segmentation mask for the target defect class (Crack).
Crack pixels must be painted white (255); all other pixels must be black (0).

All annotations, bounding boxes, labels, and numbers must be completely removed.
The output image must preserve the original resolution, geometry, and viewpoint. \\

\hline
\end{tabular}
\caption{Multi-level evaluation questions and prompt specifications used in DefectBench.}
\label{tab:prompts}
\end{table*}

\end{document}